\documentclass[runningheads]{llncs}
\usepackage[T1]{fontenc}

\usepackage{microtype}
\usepackage{makecell}
\usepackage{multirow}
\usepackage{amssymb}
\usepackage{pifont}
\usepackage[T1]{fontenc}
\usepackage{makecell}
\usepackage{wrapfig}
\renewcommand{\paragraph}[1]{\vspace{.5em}\noindent\textbf{#1}}

\usepackage{siunitx}

\usepackage[breaklinks,colorlinks]{hyperref}
\usepackage[capitalize]{cleveref}
\usepackage[dvipsnames]{xcolor}
\usepackage[table]{xcolor}
\usepackage{booktabs}
\usepackage{xspace}

\definecolor{LightGreen}{RGB}{210,245,210}
\definecolor{LightYellow}{RGB}{255,250,205}
\definecolor{LightRed}{RGB}{255,220,220}
\definecolor{LightBlue}{RGB}{220,235,255}

\makeatletter
\DeclareRobustCommand\onedot{\futurelet\@let@token\@onedot}
\def\@onedot{\ifx\@let@token.\else.\null\fi\xspace}

\def\etal{\emph{et al}\onedot}
\makeatother

\usepackage{graphicx,verbatim}

\begin{document}
\title{MIL-PF: Multiple Instance Learning on Precomputed Features for Mammography Classification}

\begin{comment}  

\end{comment}

\author{
Nikola Jovišić\inst{1,2}\thanks{Corresponding author: nikola.jovisic@ivi.ac.rs} \and
Milica Škipina\inst{1,2,3} \and
Nicola Dall'Asen\inst{3,4,5} \and
Dubravko Ćulibrk\inst{2}
}

\authorrunning{N. Jovišić et al.}

\institute{
Institute for AI R\&D of Serbia, Serbia \and
Faculty of Technical Sciences, University of Novi Sad, Serbia \and
University of Trento, Italy \and
University of Pisa, Italy \and
Fondazione Bruno Kessler, Italy
}

\maketitle
\begin{abstract}
% \begin{abstract}
Modern foundation models provide highly expressive visual representations, yet adapting them to high-resolution medical imaging remains challenging due to limited annotations and weak supervision. Mammography, in particular, is characterized by large images, variable multi-view studies and predominantly breast-level labels, making end-to-end fine-tuning computationally expensive and often impractical. We propose Multiple Instance Learning on Precomputed Features (MIL-PF), a scalable framework that combines frozen foundation encoders with a lightweight MIL head for mammography classification. By precomputing the semantic representations and training only a small task-specific aggregation module (40k parameters), the method enables efficient experimentation and adaptation without retraining large backbones. The architecture explicitly models the global tissue context and the sparse local lesion signals through attention-based aggregation. MIL-PF achieves state-of-the-art classification performance at clinical scale while substantially reducing training complexity. We release the code for full reproducibility.
% \end{abstract}


\end{abstract}

\section{Introduction}
\label{sec:intro}

Breast cancer is the most common malignancy and leading cause of cancer deaths in women \cite{ghoncheh2016incidence}. The preferred screening and diagnostic method is mammography, and the analysis of mammograms requires considerable effort from radiologists, who have been increasingly relying on computer-aided diagnosis (CAD) \cite{doi2007computer}. 

Among the clinical imaging modalities, mammography stands out in terms of high spatial resolution (up to $4708\times5844$ pixels) \cite{rangarajan2022ultra} and, like many other medical domains, typically lacks the rich, pixel-level annotations or textual supervision of a scale required for many modern machine learning training paradigms. E.g., CLIP \cite{radford2021learning}-style training is limited not only by this lack of supervision, but also by the architectural constraints regarding the high-resolution inputs.

Instead of training the encoders, we show that, even when completely frozen, the largest variant of pretrained DINOv2 \cite{oquab2023dinov2} and MedSigLIP \cite{sellergren2025medgemma} generalize outstandingly well on the out-of-distribution mammography domain.

As the experiments on the task-specific head of the network are thus conducted in a fixed representation space, the embeddings can be precomputed, massively saving on the total computation cost of the experiments. As large research laboratories continue releasing specialized encoders and the community focuses more on sustainable AI, we anticipate that this approach will become even more valuable, particularly for research groups with limited resources.

To complete the pipeline, we frame the task using \textbf{Multiple Instance Learning (MIL)}. This allows us to adapt to realistic clinical scenarios where a single label is associated with a ``bag'' of instances (all views of a single breast). We explore this task in light of its hierarchical MIL structure, multiscale signal importance, and weakly-labeled regime. The architecture we propose for this task sits on top of the pretrained foundational encoder and consists of only $\sim40k$ trainable parameters, making clinical adoption easy.

We evaluate MIL-PF on the mammography breast-level classification task (with visual explainability), but formalize a framework to make domain generalization accessible. Our main contributions are as follows: 1. We formalize a class of mammography-motivated MIL problems and propose an architecture designed to address them. 2. We leverage recent advances in foundation vision models to revisit feature precomputing and frozen encoders as a principled design choice for MIL, showing that strong general backbones enable efficient and expressive pipelines without the need for end-to-end fine-tuning. 3. We validate the approach at clinical scale, demonstrate state-of-the-art performance on mammography classification benchmarks and provide code to enable full reproducibility.
\section{Method}
\label{sec:method}

\begin{figure*}[th]
    \centering
    \includegraphics[width=\textwidth]{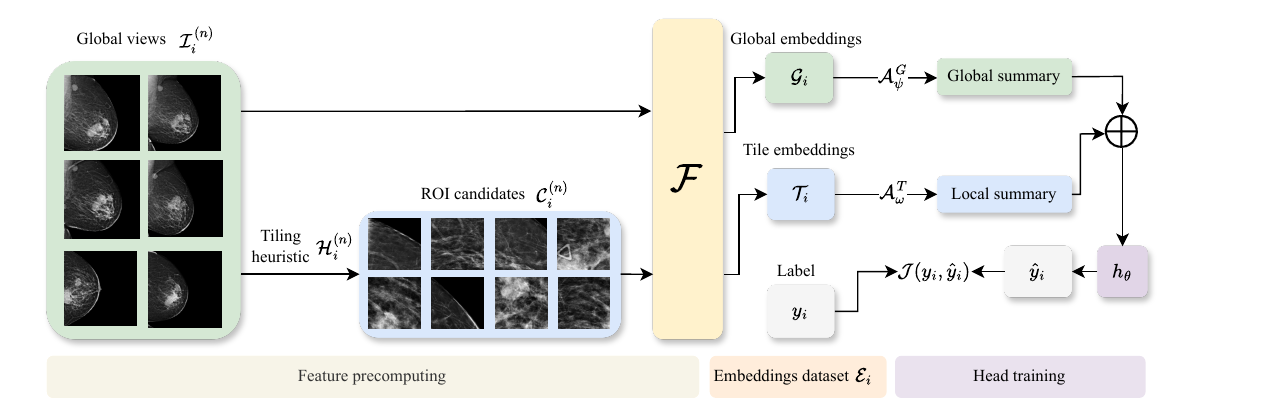}
    \caption{The MIL-PF pipeline diagram, divided into distinct stages: Feature Precomputing producing Embeddings dataset $\mathcal{E}$ and Head Training using them.
    }
    \label{fig:method}
\end{figure*}

\paragraph{Multiple Instance Learning (MIL).}
In Multiple Instance Learning (MIL) \cite{dietterich1997solving,zaheer2017deep,ilse2018attention}, the dataset $\mathcal{D}$ is composed of $T$ labeled bags, as defined in \cref{eq:dataset}.

\begin{equation}
\mathcal{D} = \left\{ (\mathcal{S}_i, y_i) \right\}_{i=1}^{T}, 
\quad \text{where } 
\mathcal{S}_i = \{ I_i^{(n)} \}_{n=1}^{N_i} 
\text{ and } y_i \in \mathcal{L}
\label{eq:dataset}
\end{equation}

Each bag $\mathcal{S}_i$ is a collection of $N_i$ instances $I_i^{(n)}$. The label $y_i \in \mathcal{L}$ is a bag-level label that applies to the entire bag $\mathcal{S}_i$; the labels for individual instances $I_i^{(n)}$ are unknown. 
The learning task is to map each bag $\mathcal{S}_i$ to its corresponding label $y_i$. 

\paragraph{Problem formulation.}
We study a special case of MIL, which includes:
\begin{enumerate}
    \item \textbf{Nested hierarchy:} Each instance $I_i^{(n)}$ from a bag $\mathcal{S}_i$ is a high-resolution image. This image, in turn, contains a variable number of sparsely distributed Regions of Interest (ROIs), $\{R_i^{(n)(k)}\}$. Crucially, these ROIs are not annotated.
    \item \textbf{Complementary streams:} The images $\{I_i^{(n)}\}_n$ within a bag convey a comparable low-frequency signal (global stream), e.g. pertinent to general tissue structure, while substantial, fine-grained differences exist only at high resolution within their respective ROIs (local stream). 
\end{enumerate}

\paragraph{Embeddings dataset.}
MIL methods have traditionally been shown to have problems with backpropagation, especially with sparsely present task-relevant instances \cite{ilse2018attention}. Multiscale reasoning, as required by our problem, hinders it further.

Therefore, we first transform our data using a pretrained encoder. We assume the existence of an encoder, $\mathcal{F}$, that is pretrained to expressively encode both full images $I_i^{(n)}$ and high-resolution sub-regions, the tiles, in a zero-shot setting. Using this encoder, we generate two distinct sets of embeddings for each bag $\mathcal{S}_i$.

First, we create a set of global embeddings, $\mathcal{G}_i$, by applying the encoder to each full image in the bag, as in \cref{eq:global}.
\begin{equation}
\mathcal{G}_i = \{\mathcal{F}(I_i^{(n)}) \}_n
\label{eq:global}
\end{equation}

Second, to capture the local signals, we must extract features from the unannotated ROIs. We create candidates for these ROIs, $\{C_i^{(n)(k)}\}_{k=1}^{M_i^{(n)}}$, by dividing each image $I_i^{(n)}$ into a uniform grid of tiles. The heuristic $\mathcal{H}_i^{(n)}$ we use for selecting the tiles is taking all containing any breast tissue from the grid of encoder-input-sized tiles, discarding only the fully background ones. The tile size must be large enough to fit the expected ROI, but not larger than the maximum resolution the encoder supports. The number of tiles $M_i^{(n)}$ may vary per image.

We then create a single set of tile embeddings, $\mathcal{T}_i$, by encoding all candidate tiles from all images in the bag and collecting them together, as in \cref{eq:local}.
\begin{equation}
\mathcal{T}_i = 
\bigcup_{n} \bigcup_{k=1}^{M_i^{(n)}} 
\{ \mathcal{F}(C_i^{(n)(k)}) \}
\label{eq:local}
\end{equation}

Finally, as defined in \cref{eq:unified}, we form a unified embeddings dataset, $\mathcal{E}$, which consists of the global embeddings, the tile embeddings, and the corresponding bag label $y_i$ for each bag in our original dataset.
\begin{equation}
\mathcal{E} =
\left\{
\left(
\mathcal{G}_i,
\mathcal{T}_i, y_i
\right)
\right\}_i
\label{eq:unified}
\end{equation}

\paragraph{MIL-PF Head.}
For simplicity, our model employs a late-fusion strategy of the two streams, as on \cref{fig:method}. Even though learning complex interactions of the global $\mathcal{G}_i$ and the local $\mathcal{T}_i$ stream could in principle be useful, it is not crucial for our task. We define inference in \cref{eq:inference}.
\begin{equation}
\hat{y}_i
=
h_\theta \!\left(
concat
\left(
\mathcal{A}^{G}_{\psi}\!\left( \mathcal{G}_i \right),
\mathcal{A}^{T}_{\omega}\!\left( \mathcal{T}_i \right)
\right)
\right),
\quad
\left(
\mathcal{G}_i, \mathcal{T}_i, y_i
\right) \in \mathcal{E}
\label{eq:inference}
\end{equation}

\begin{itemize}
    \item $\mathcal{A}^{G}_{\psi}$ and $\mathcal{A}^{T}_{\omega}$ are permutation-invariant aggregators with parameters $\psi$ and $\omega$ that produce a single summary vector for the global and local embedding sets, respectively. Prior to actual pooling operations, they pass the embeddings through a small Multi-Layer Perceptron (MLP). As the encoder is frozen, this part does the actual processing of individual crops with respect to the task, and this is where the majority of the trainable parameters are.
    \item $h_{\theta}$ is a final layer that maps the joint summary to a prediction $\hat{y}_i$.
\end{itemize}

The training objective minimizes the loss $\mathcal{J}(y_i, \hat{y}_i)$ with respect to  $(\theta, \psi, \omega)$.

A key advantage of this design is that the pretrained encoder $\mathcal{F}$ performs most of the heavy lifting. The MIL-PF module itself can therefore be lightweight with a low parameter count, focusing solely on correctly implementing the domain's inductive biases. As shown in \cref{sec:experiments_and_results}, this modularity provides numerous benefits: fast experiment iteration, drastically lower computational requirements and the ease of swapping encoders or inserting additional inductive biases.

\paragraph{Attention Pooling for Local Embeddings.}
A careful choice of the local aggregator ($\mathcal{A}^{T}_{\omega}$) is necessary, as the task-relevant ROIs are sparse in the set $\mathcal{T}_i$.

Standard pooling methods are suboptimal for this task. Mean pooling fails, as the signal from a few important tiles is suppressed or ``diluted'' by the overwhelming number of unimportant background tiles, and max pooling can only capture information from a single, most-salient tile, even if multiple ROIs are relevant for the bag-level prediction. An attention mechanism is therefore preferred. However, common candidates like self-attention are ill-suited, as they are designed to primarily model the \textit{interdependence} between inputs, while we are interested just in attending to relevant tiles.

Therefore, we employ a \textbf{Perceiver-style cross-attention} mechanism \cite{jaegle2021perceiver} as our local aggregator $\mathcal{A}^{T}_{\omega}$. This approach uses a single, trainable latent vector $z$ that acts as a query. This latent query $z$ attends to the set of all local tile embeddings from $\mathcal{T}_i$, which are projected into Keys (K) and Values (V). The mechanism computes a weighted sum of the Values based on the query-key similarity ($\text{softmax}(z K^T)V$), which ``pulls'' the most relevant information from the tile set into a single summary vector.

This design is highly parameter-efficient, focusing the model's capacity on learning the weights of the aggregators and the final head. In our experiments, we found that using a single latent query was sufficient and that adding more latents (as in the standard Perceiver) did not provide additional benefits for this task.

\section{Experiments and Results}
\label{sec:experiments_and_results}

\subsection{Experimental Setup}

\paragraph{Datasets.}  
We evaluate on one of the largest public mammography datasets, the EMBED Open Data, which contains data for $\sim0.5$M mammograms \cite{jeong2023emory}, representing highly diverse real-world clinical scenarios  \cite{woo2025subgroup}. 
For comparison with other mammography approaches, we also evaluate on VinDr \cite{nguyen2023vindr} and RSNA \cite{rsna}.

A bag $\mathcal{S}_i$ is defined as a set of all views of a single breast in a given exam. We use breast-wise labels, as opposed to image-wise, because radiologists form reports considering all of the views together. Patient-wise reasoning, on the other hand, collapses ("max-pools") to the label of the more suspicious breast, losing valuable training signal. Lesions, defined as tissue changes that could be related to breast cancer, represent our $R_i$. Annotating ROIs is very hard to scale due to the cost and highly diverse pathologies. 

Even the weak case-level label assignment strategy (BI-RADS reports) is noisy and varies among radiologists. Therefore, for general malignancy risk modeling evaluation, we use BI-RADS 1 as negative, and BI-RADS 4, 5 and 6 as positive, across mixed clinical cohorts. The split choice varies widely in literature and this specific one minimizes noise and assures fair statistical comparison, while retaining principal clinical significance. Using solely BI-RADS 6 as the only actual ground truth is statistically unreliable, since those examples are scarce even in the largest datasets.

\paragraph{Metrics.} The metrics used in the literature vary. We use \textbf{AUC} for class separability; \textbf{bACC} to reflect performance regarding class inbalance; \textbf{mAP} for lesion detection; and \textbf{Spec@Sens=0.9}, specificity at a threshold that yields a sensitivity of 0.9. We chose the last metric due to clinical relevance of sensitivity \cite{petitti2005saving}; one is often concerned about how much specificity can be achieved with a high sensitivity constraint. We note it does not strictly correlate with AUC.

\paragraph{Implementation Details.} We split the data 70\%-10\%-20\% for train, validation and test sets, balanced for all used BI-RADS values proportionally, assuring no patient leakage across splits. The MLP part of the $\mathcal{A}^G$ and $\mathcal{A}^T$ consist of two layers followed by ReLU activations, the first layer projecting from embedding size to 16, and second from 16 to 8. The loss function used is Binary Cross-Entropy. For every experiment, all bags of each split of the whole embeddings dataset $\mathcal{E}$, even for the large EMBED dataset, can fit in a single batch of our A100 40GB GPU and we train in such a setting, requiring $\sim$ 5-7 min for a single run of full MIL-PF head training and $\sim$ 2M FLOPS for a forward pass per breast. For classification, we do not use overlapping tiles, as we observed no improvement in performance. For the attention map computation we use 75\% overlap at inference time. IoU threshold for mAP is 0.25.

\paragraph{Mutliple runs.} Since the head training is inexpensive with the precomputed embeddings, we run 36 independent runs per each head training experiment. Although we observe some variance across runs (up to 2\% AUC and 11\% Spec@Sens=0.9), selecting the model with the highest validation AUC consistently yields near-maximal test performance. Thus, the results reflect a reliable optimization effect rather than a single fortunate run.

\subsection{Encoder Selection}

Training a simplified version of MIL-PF on EMBED BI-RADS task that relies just on max-pooling-based $\mathcal{A}^G$ gives us a general overview of the backbone choices. As shown in \cref{tab:two_tables}. DINOv2 \cite{oquab2023dinov2} and MedSigLIP \cite{sellergren2025medgemma} overperform  other medical and general encoders, as well as the mammography-pretrained MammoCLIP \cite{ghosh2024mammo}, which does not generalize as well to a new dataset.

\subsection{Comparison with the baselines}

We compare MIL-PF against recent state-of-the-art models that address a similar task and have publicly available code for reproduction: Shen \etal \cite{shen2021interpretable}, Pathak \etal \cite{pathak2025breast} and Mour{\~a}o \etal \cite{mourao2025multi}. As the competing models are not breast-level-aligned, we additionally evaluate the main competitor (Mour{\~a}o \etal) using mean and max aggregation over multiple views. All the models were retrained on the same data splits as MIL-PF on VinDr, RSNA and EMBED for fair comparison.

\cref{tab:baselines} and \cref{tab:mass_calc} show MIL-PF's competitive or superior performance, especially on the largest scale and most diverse EMBED dataset. We do not attribute this advantage to any single component of our method by itself, but to a combination of factors: MIL-PF being the first in the literature to discover generalization properties of the general encoders to the mammography domain; this further enabling feature precomputing, fast iterative experimentation and multiple runs optimization; and finally resulting in a domain-aligned aggregation mechanism (ablated in \cref{sec:ablation}) with inherent breast-wise modeling.
\begin{table}[!ht]
\centering
%\caption{Quantitative comparison of proposed MIL-PF \wrt SOTA models. On the more complete and challenging dataset EMBED, MIL-PF shows superior performance than all previous methods.}
\caption{Quantitative comparison of MIL-PF against SOTA baselines on VinDr and EMBED on BI-RADS-based malignancy. MIL-PF demonstrates superior performance, particularly on the large-scale, noisy EMBED dataset, highlighting its robustness.}
{\resizebox{\columnwidth}{!}{
\begin{tabular}{lrlcccccc}
\toprule
& & & \multicolumn{3}{c}{\textbf{VinDr}}  & \multicolumn{3}{c}{\textbf{EMBED}} \\ 
\cmidrule(lr){4-6}\cmidrule(lr){7-9}
\textbf{Model/Paper}& \makecell{\textbf{\# Trainable} \\ \textbf{ Params (M)}} & \textbf{Level} & \textbf{AUC} $\uparrow$ &  \textbf{bAcc} $\uparrow$ & \makecell{\textbf{Spec@} \\ \textbf{Sens=0.9}} $\uparrow$ & \textbf{AUC} $\uparrow$ & \textbf{bAcc} $\uparrow$ & \makecell{\textbf{Spec@} \\ \textbf{Sens=0.9}} $\uparrow$\\ 
\midrule
\rowcolor{gray!10}ES-Att$^{side}$ \cite{pathak2025breast}                 &  22.89   & Case   & 0.786             & 0.749 & 0.193             & 0.836 &  0.668 & 0.558 \\
\rowcolor{gray!10}SIL$^{IL}$ GMIC-ResNet18 \cite{pathak2025breast}               &  22.49   & Image  & \underline{0.911} & \textbf{0.855} & 0.698             & 0.875 &  0.802 & 0.566\\
\rowcolor{gray!10}GMIC \cite{shen2021interpretable}                              &  14.11   & Image  & 0.899             & 0.826 & 0.690             & 0.816 &  0.736 & 0.380\\
\rowcolor{gray!10}FPN-AbMIL \cite{mourao2025multi}                               &  1.76    & Image  & \textbf{0.920}    & 0.844 & \underline{0.720} & 0.802 &  0.740 & 0.367 \\
\rowcolor{gray!10}FPN-SetTrans \cite{mourao2025multi}                            &  5.38    & Image  & 0.908             & 0.827 & 0.670             & 0.794 &  0.728 & 0.364 \\

\rowcolor{gray!10}FPN-AbMIL (mean) \cite{mourao2025multi}                     & 1.76  & Breast & \underline{0.911} & 0.831 & 0.643 & 0.835 & 0.765 & 0.403 \\
\rowcolor{gray!10}FPN-AbMIL (max) \cite{mourao2025multi}                      & 5.38  & Breast & 0.899 & 0.830 & 0.512 & 0.817 & 0.759 & 0.334 \\
\rowcolor{gray!10}FPN-SetTrans (mean) \cite{mourao2025multi}                  & 1.76  & Breast & 0.900 & 0.811 & 0.559 & 0.835 & 0.767 & 0.451 \\
\rowcolor{gray!10}FPN-SetTrans (max) \cite{mourao2025multi}                   & 5.38  & Breast & 0.887 & 0.776 & 0.595 & 0.813 & 0.728 & 0.388  \\

\rowcolor{ForestGreen!10}\textbf{MIL-PF (Ours, DINOv2)} $\mathcal{A}^T$: max        &  0.05    & Breast & 0.882             & 0.811 & 0.712             & 0.905 &  0.833 & 0.703\\
\rowcolor{ForestGreen!10}\textbf{MIL-PF (Ours, DINOv2)} $\mathcal{A}^T$: attn       &  0.05    & Breast & 0.894             & 0.845 & \textbf{0.792}    & \underline{0.916} & \textbf{0.850} & \textbf{0.762}  \\
\rowcolor{ForestGreen!10}\textbf{MIL-PF (Ours, MedSigLIP)} $\mathcal{A}^T$: max     &  0.04    & Breast & 0.901             & 0.838 & 0.616             & \textbf{0.918}  & \underline{0.845}  & 0.735 \\
\rowcolor{ForestGreen!10}\textbf{MIL-PF (Ours, MedSigLIP)} $\mathcal{A}^T$: attn    &  0.04    & Breast & \underline{0.911} & \underline{0.850} & 0.683             & 0.914 &    0.843    & \underline{0.746}   \\
\bottomrule
\end{tabular}
}}
\label{tab:baselines}
\end{table}

\begin{table}[!ht]
\centering
%\caption{Quantitative comparison of proposed MIL-PF \wrt SOTA models. On the more complete and challenging dataset EMBED, MIL-PF shows superior performance than all previous methods.}
\caption{Quantitative comparison of MIL-PF against SOTA baselines on the VinDr and RSNA datasets. MIL-PF demonstrates superior performance.}
{\resizebox{\columnwidth}{!}{
\begin{tabular}{lrlccccccccc}
\toprule
& & & \multicolumn{3}{c}{\textbf{VinDr - Mass}}  & \multicolumn{3}{c}{\textbf{VinDr - Calcification}} & \multicolumn{3}{c}{\textbf{RSNA - Cancer}} \\ 
\cmidrule(lr){4-6}\cmidrule(lr){7-9} \cmidrule(lr){10-12} 
\textbf{Model/Paper}& \makecell{\textbf{\# Trainable} \\ \textbf{ Params (M)}} & \textbf{Level} & \textbf{AUC} $\uparrow$ &  \textbf{bAcc} $\uparrow$ & \makecell{\textbf{Spec@} \\ \textbf{Sens=0.9}} $\uparrow$ & \textbf{AUC} $\uparrow$ & \textbf{bAcc} $\uparrow$ & \makecell{\textbf{Spec@} \\ \textbf{Sens=0.9}} $\uparrow$ & \textbf{AUC}  $\uparrow$ & \textbf{bAcc} $\uparrow$ & \makecell{\textbf{Spec@} \\ \textbf{Sens=0.9}} $\uparrow$ \\ 
\midrule
\rowcolor{gray!10}ES-Att$^{side}$ \cite{pathak2025breast}              & 22.89 & Case   & 0.738 & 0.669 & 0.266 & 0.937 & 0.864 & 0.877 & 0.890 & 0.799 & 0.642 \\
\rowcolor{gray!10}SIL$^{IL}$ GMIC-ResNet18 \cite{pathak2025breast}            & 22.49 & Image  & 0.772 & 0.703 & 0.307 & 0.953 & 0.886 & 0.886 & 0.899 & 0.754 & 0.684 \\
\rowcolor{gray!10}GMIC \cite{shen2021interpretable}                           & 14.11 & Image  & 0.755 & 0.687 & 0.286 & 0.941 & 0.855 & 0.899 & 0.773 & 0.699 & 0.434 \\
\rowcolor{gray!10}FPN-AbMIL \cite{mourao2025multi}                            & 1.76  & Image  & 0.792 & 0.720 & 0.382 & 0.954 & 0.902 & 0.904 & 0.891 & 0.793 & 0.628 \\
\rowcolor{gray!10}FPN-SetTrans \cite{mourao2025multi}                         & 5.38  & Image  & 0.785 & 0.734 & 0.367 & 0.942 & 0.884 & 0.836 & 0.874 & 0.773 & 0.656\\

\rowcolor{gray!10}FPN-AbMIL (mean) \cite{mourao2025multi}                 & 1.76  & Breast & \underline{0.808} & \textbf{0.757} & 0.387 & 0.962 & 0.913 & 0.916 & 0.914 & 0.826 & 0.687 \\
\rowcolor{gray!10}FPN-AbMIL (max) \cite{mourao2025multi}                  & 5.38  & Breast & 0.797 & 0.683 & 0.332 & \underline{0.958} & 0.899 & 0.911 & 0.902 & 0.785 & 0.664 \\
\rowcolor{gray!10}FPN-SetTrans (mean) \cite{mourao2025multi}              & 1.76  & Breast & 0.807 & \underline{0.755} & 0.317 & 0.952 & 0.881 & 0.867 & 0.902 & 0.825 & \underline{0.747} \\
\rowcolor{gray!10}FPN-SetTrans (max) \cite{mourao2025multi}               & 5.38  & Breast & 0.801 & 0.745 & 0.342 & 0.947 & 0.877 & 0.892 & 0.896 & 0.773 & 0.693 \\

\rowcolor{ForestGreen!10}\textbf{MIL-PF (Ours, DINOv2)} $\mathcal{A}^T$: max     & 0.05  & Breast & 0.770 & 0.731 & 0.311 & \underline{0.958} & \underline{0.917} & 0.884 & 0.901 & 0.815 & 0.669 \\
\rowcolor{ForestGreen!10}\textbf{MIL-PF (Ours, DINOv2)} $\mathcal{A}^T$: attn    & 0.05  & Breast & 0.800 & 0.736 & \textbf{0.419} & \textbf{0.967} & \textbf{0.930} & \textbf{0.931} & \underline{0.923} & 0.828 & 0.709 \\
\rowcolor{ForestGreen!10}\textbf{MIL-PF (Ours, MedSigLIP)} $\mathcal{A}^T$: max  & 0.04  & Breast & 0.778 & 0.720 & \underline{0.408} & \textbf{0.967} & 0.907 & 0.861 & \textbf{0.925} & \underline{0.830} & 0.733  \\
\rowcolor{ForestGreen!10}\textbf{MIL-PF (Ours, MedSigLIP)} $\mathcal{A}^T$: attn & 0.04  & Breast & \textbf{0.814} & \textbf{0.757} & 0.403 & \textbf{0.967} & 0.835 & \underline{0.917} & \underline{0.923} & \textbf{0.835} & \textbf{0.757}  \\
\bottomrule
\end{tabular}
}}
\label{tab:mass_calc}
\end{table}

% \begin{table}[htpb]
% \centering
% \resizebox{\columnwidth}{!}{
% \begin{tabular}{llc}
% \toprule
% \textbf{Model} & \textbf{AUC} \\
% \midrule
% \multicolumn{2}{l}{\textbf{VinDr}} \\
% \midrule
% % Yang \etal (2024) \cite{yang2024mammo} & 0.828 \\
% % Kebede \etal (2024) \cite{kebede2024dual} & 0.830 \\
% GMIC & \\
% ES-Att$^{side}$ \cite{pathak2025breast} \cite{pathak2025breast} & 0.830 \\
% SIL$^{IL}$ GMIC-ResNet18 \cite{pathak2025breast}  & 0.911 \\
% % Ibragimov \etal (2024) \cite{ibragimov2024mamt} & 0.840 \\
% % Petrini \etal (2025) \cite{petrini2025optimizing} & 0.851 \\
% % Lamprou \etal (2025) \cite{lamprou2024stethonet} & 0.857 \\
% (ours) APMF (DINOv2 ViT Giant) $\mathcal{A}^T$: max & 0.870 \\
% (ours) APMF (DINOv2 ViT Giant) $\mathcal{A}^T$: attn & 0.894 \\
% (ours) APMF (MedSigLIP) $\mathcal{A}^T$: max & \underline{0.895} \\
% (ours) APMF (MedSigLIP) $\mathcal{A}^T$: attn & \textbf{0.907} \\
% \midrule

% \multicolumn{2}{l}{\textbf{EMBED}} \\
% \midrule
% Park \etal (2025) \cite{park2025multi} & 0.890\\
% SIL$^{IL}$ GMIC-ResNet18 \cite{pathak2025breast}  & 0.790\\
% ES-Att$^{side}$ \cite{pathak2025breast} & 0.696 \\

% (ours) APMF (DINOv2 ViT Giant) $\mathcal{A}^T$: max & 0.905 \\
% (ours) APMF (DINOv2 ViT Giant) $\mathcal{A}^T$: attn & \underline{0.916} \\
% (ours) APMF (MedSigLIP) $\mathcal{A}^T$: max & \textbf{0.918} \\
% (ours) APMF (MedSigLIP) $\mathcal{A}^T$: attn & 0.914 \\
% \bottomrule
% \end{tabular}}
% \caption{.}
% \label{tab:baselines}
% \end{table}
% \smallskip
% \textit{*Significant at $p < 0.05$.}

\subsection{Explainability Analysis}

\begin{figure*}[t]
\centering

\setlength{\tabcolsep}{1pt} % horizontal spacing
\renewcommand{\arraystretch}{1} % vertical spacing

\begin{tabular}{c c c c c c c}

% -------- Row 1 --------
\rotatebox{90}{\textbf{Calcification}} &
\includegraphics[width=0.15\textwidth]{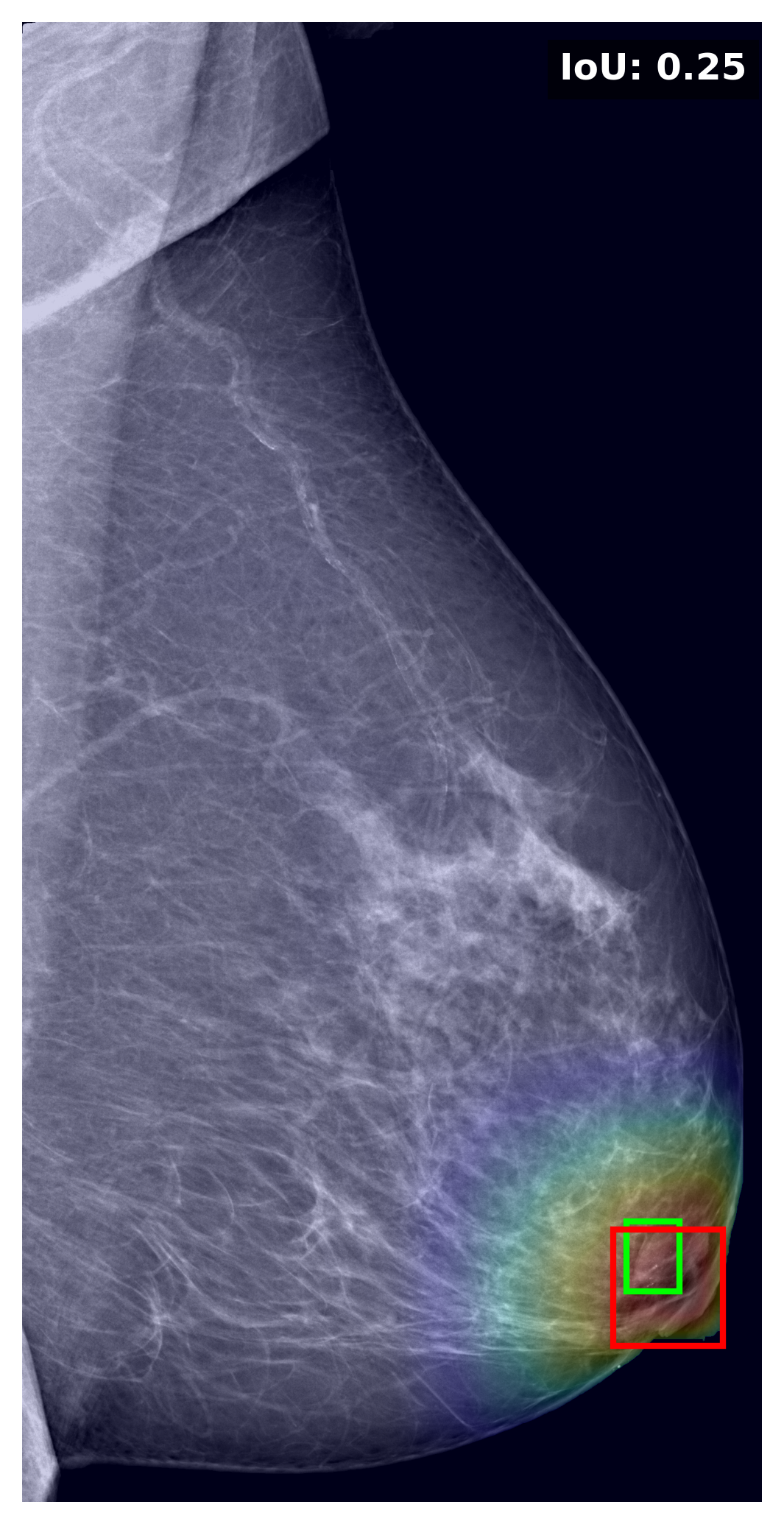} &
\includegraphics[width=0.15\textwidth]{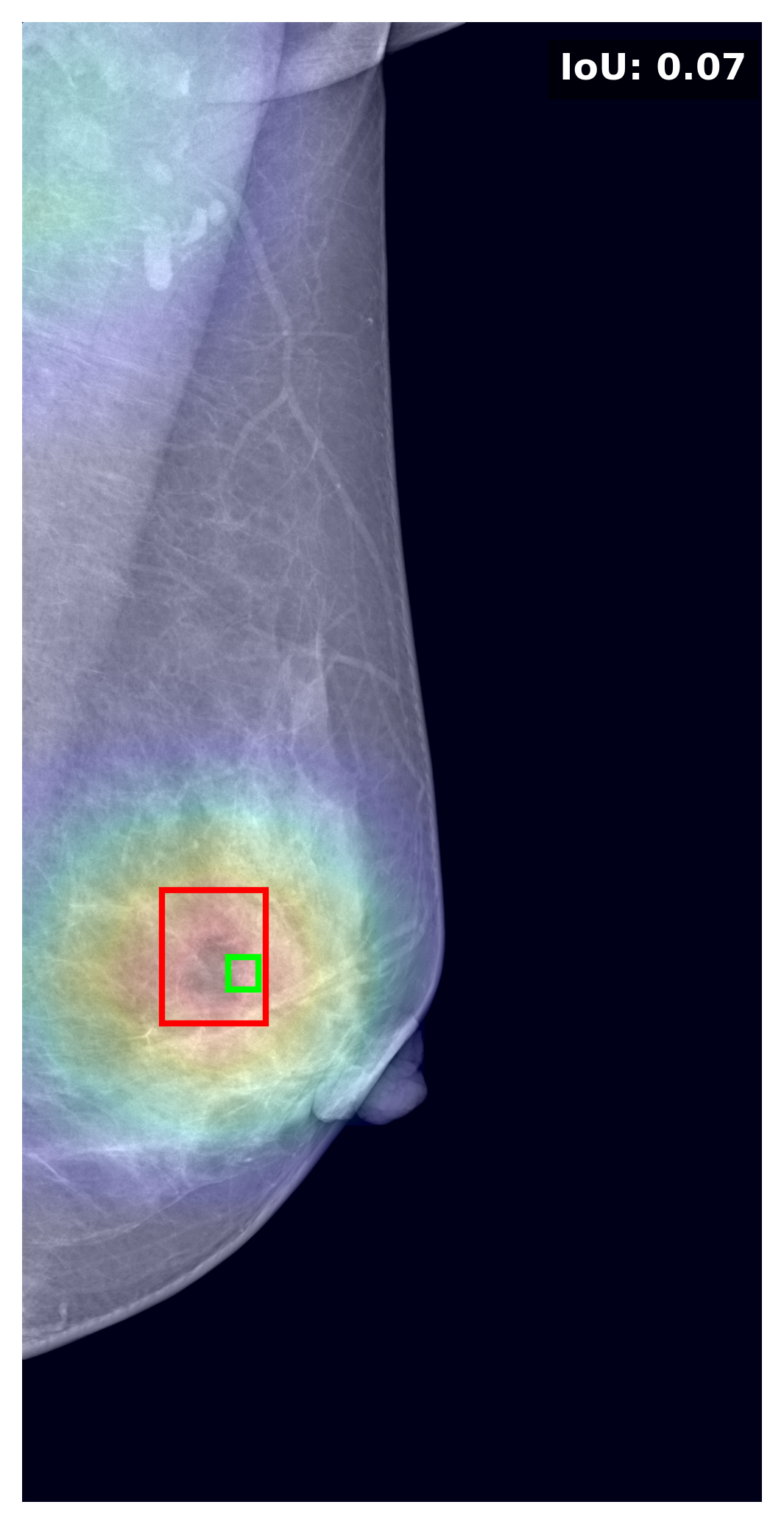} &
\includegraphics[width=0.15\textwidth]{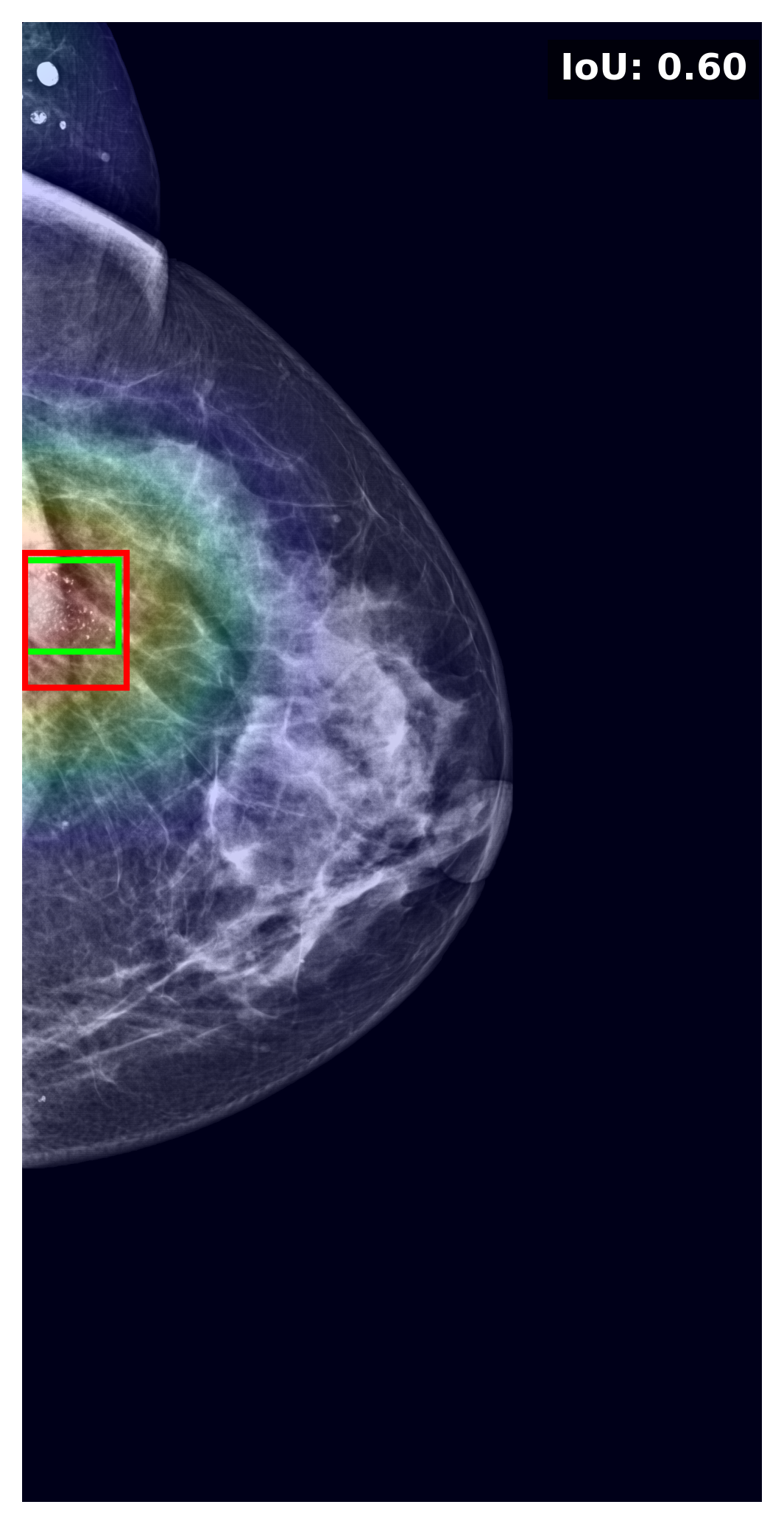} &
\includegraphics[width=0.15\textwidth]{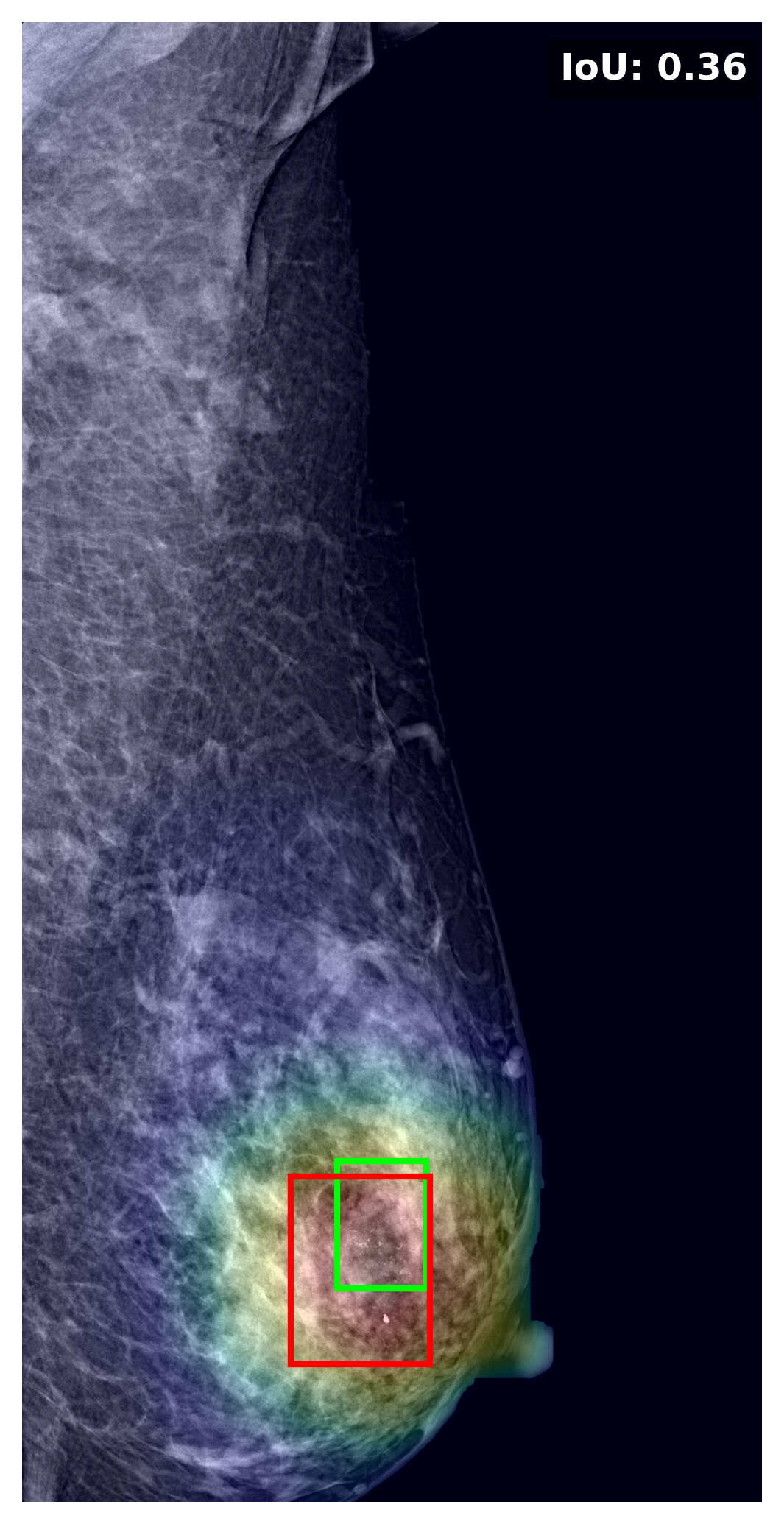} &
\includegraphics[width=0.15\textwidth]{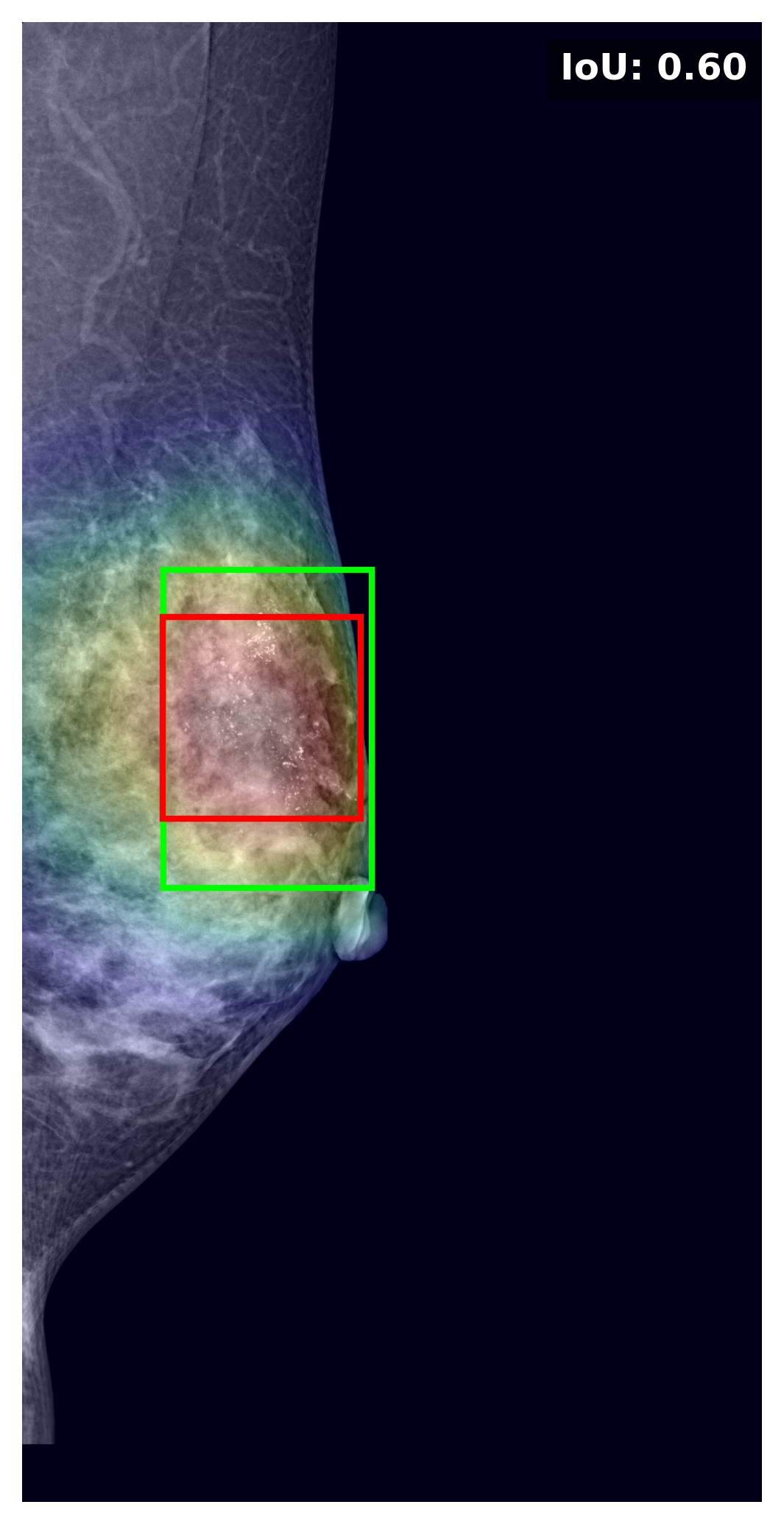} &
\includegraphics[width=0.15\textwidth]{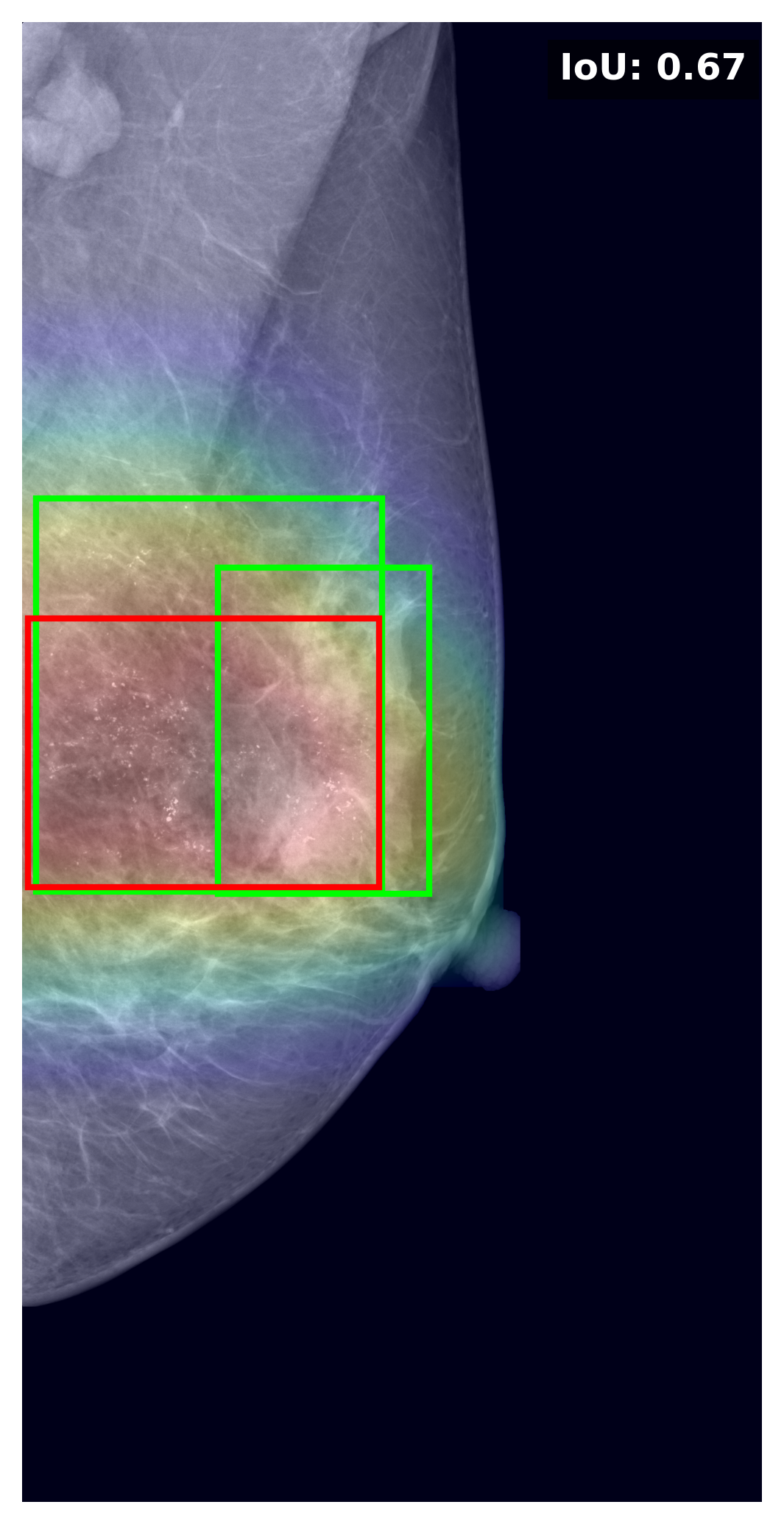} \\

% -------- Row 2 --------
\rotatebox{90}{\textbf{Mass}} &
\includegraphics[width=0.15\textwidth]{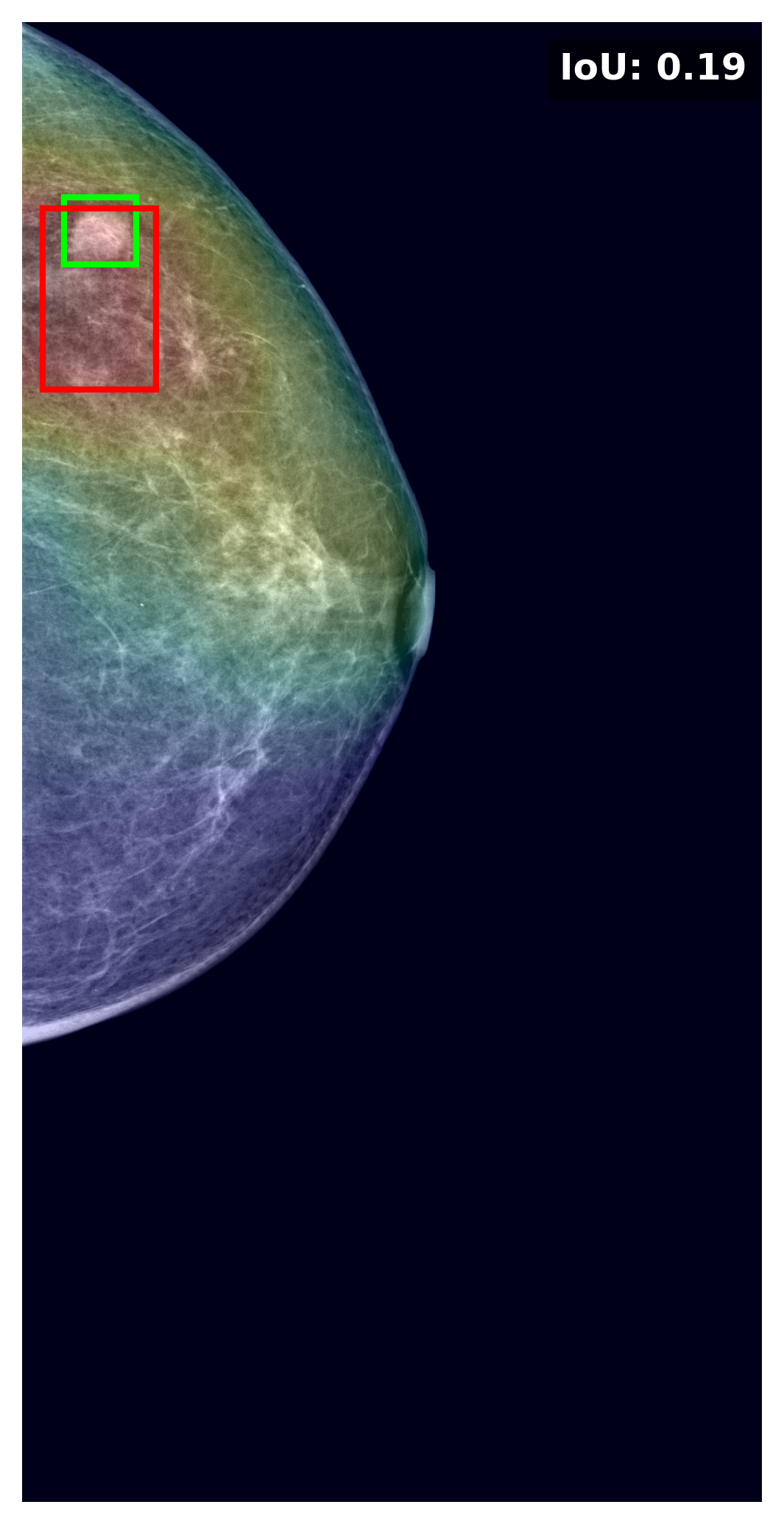} &
\includegraphics[width=0.15\textwidth]{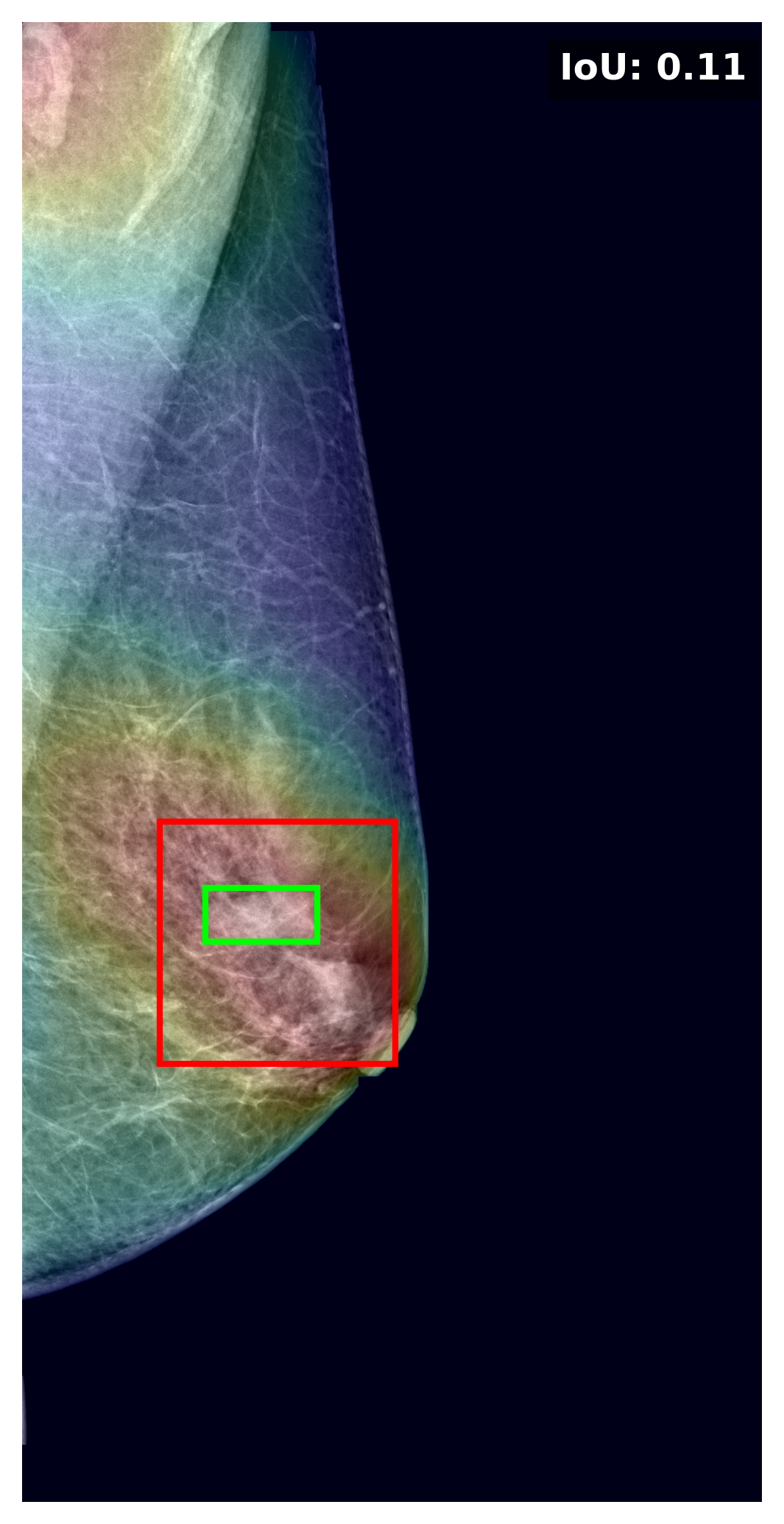} &
\includegraphics[width=0.15\textwidth]{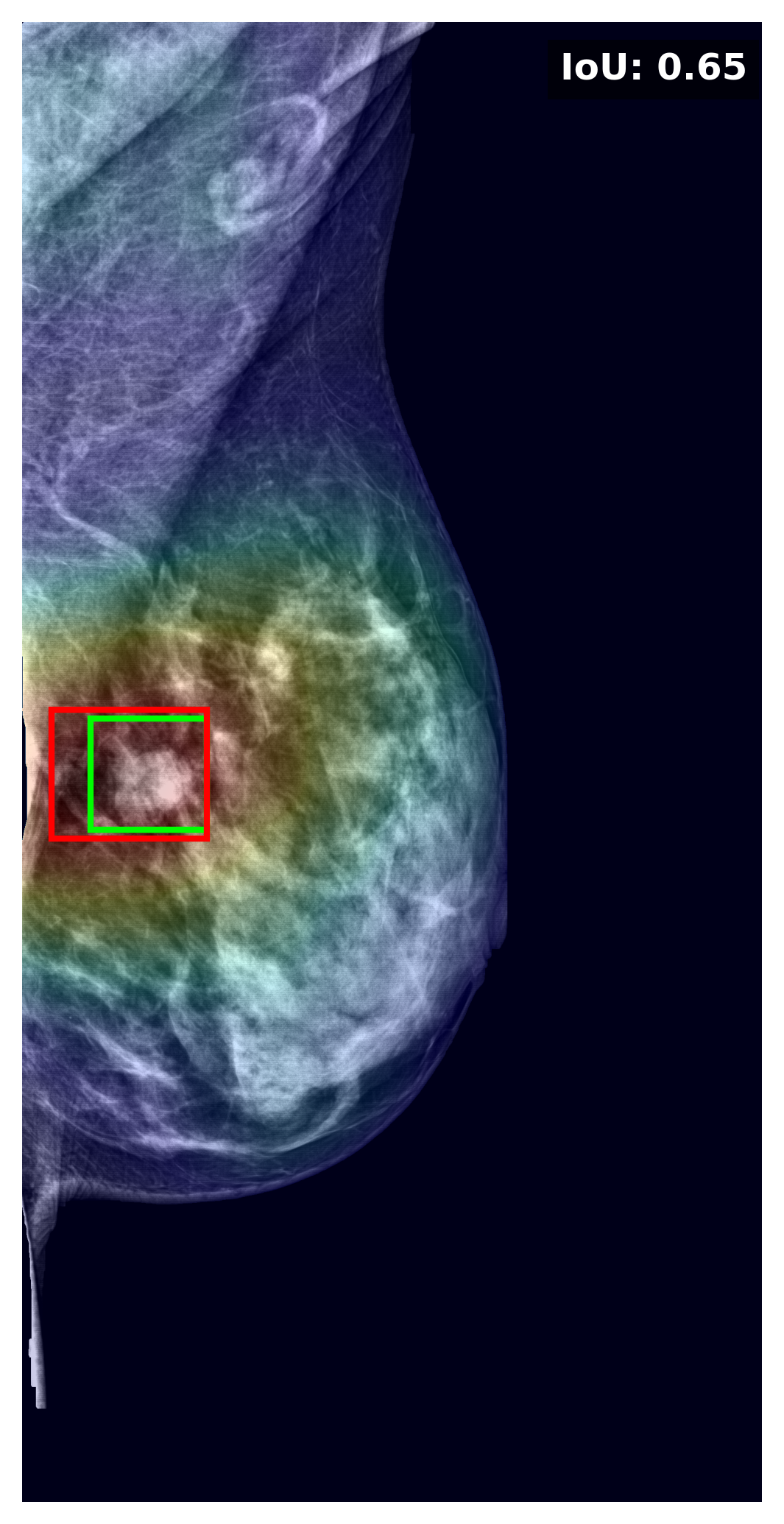} &
\includegraphics[width=0.15\textwidth]{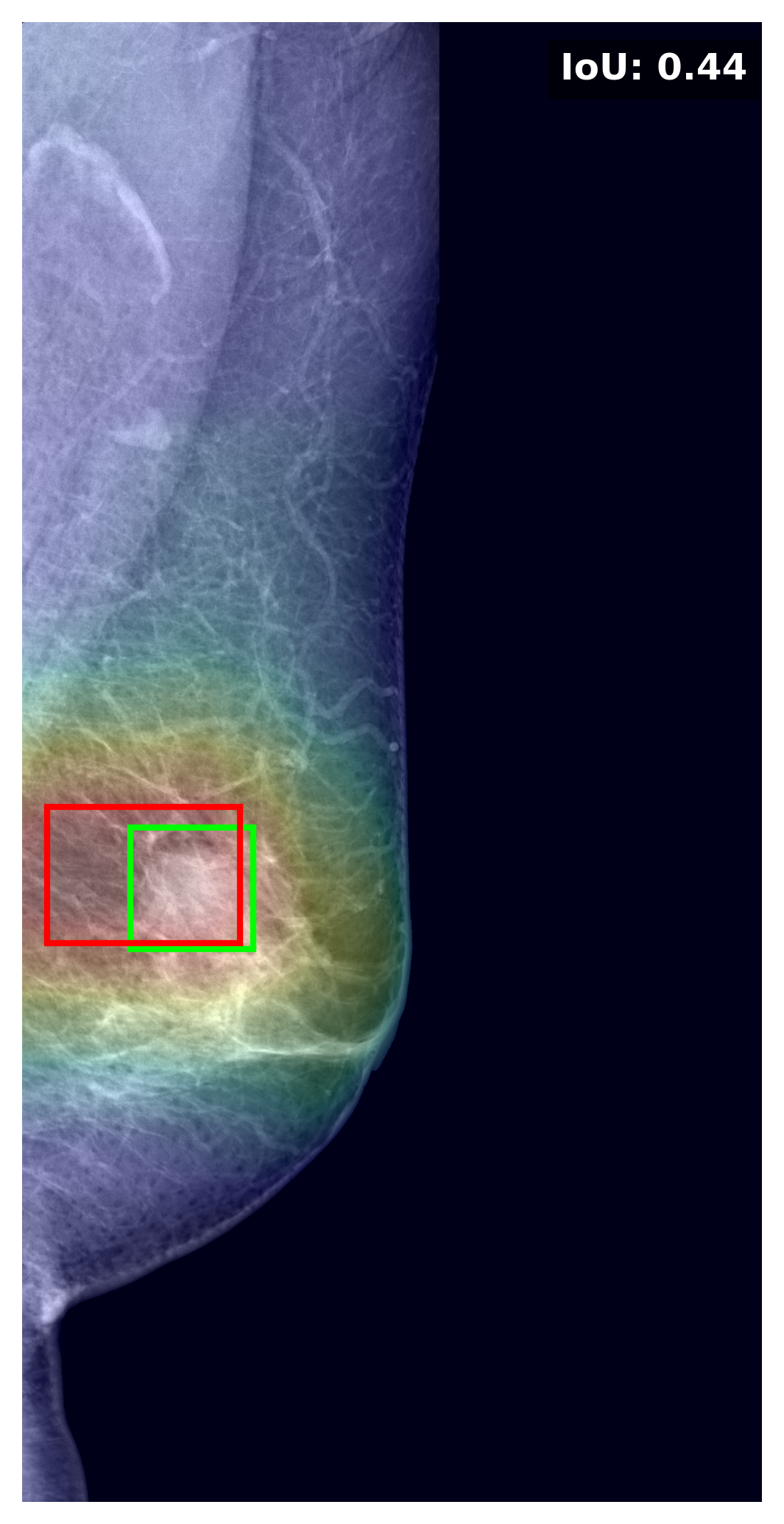} &
\includegraphics[width=0.15\textwidth]{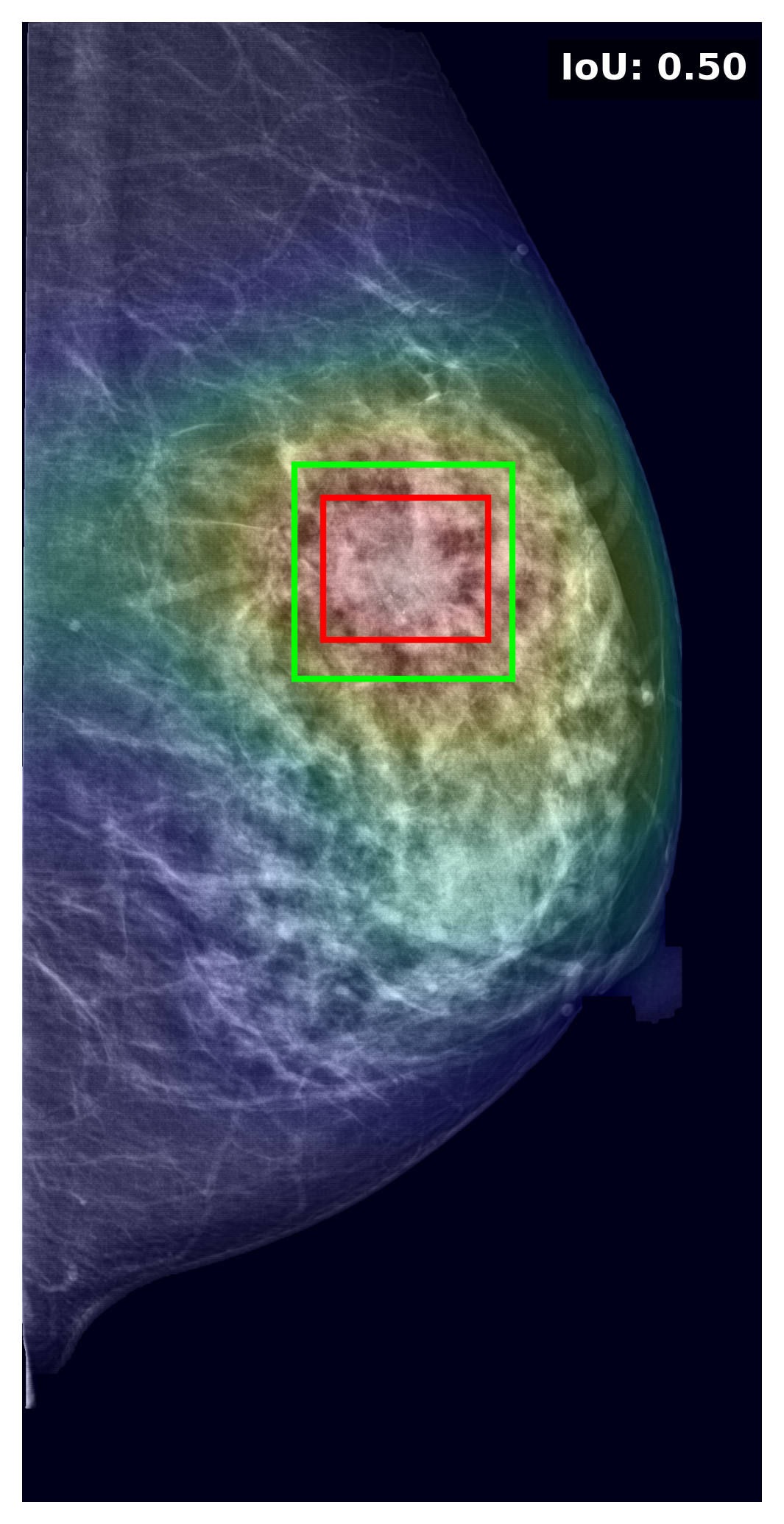} &
\includegraphics[width=0.15\textwidth]{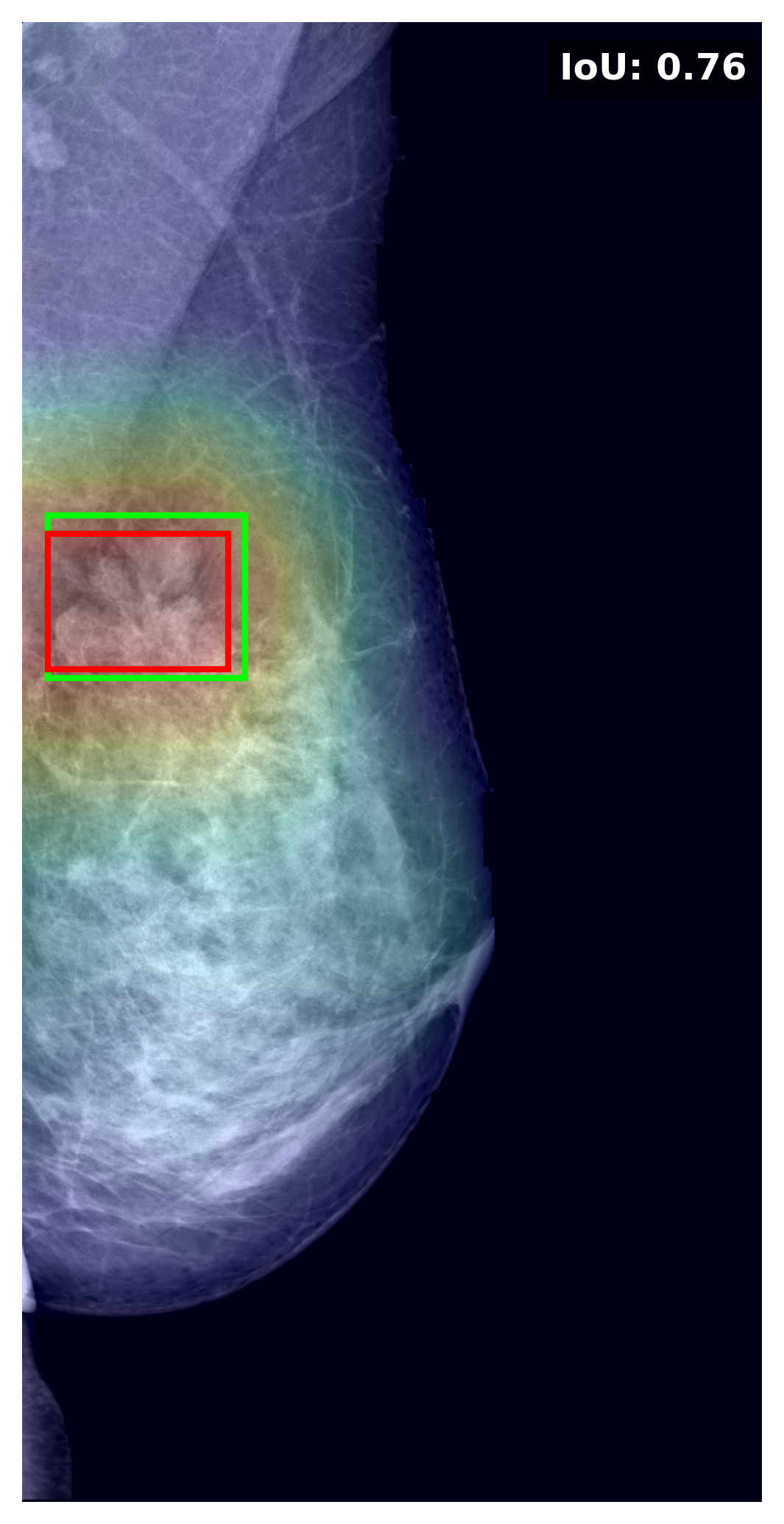} \\

\end{tabular}

\caption{Attention maps for masses and calcifications obtained by overlapping the MIL-PF's local stream window. Ground truth is shown in green and predictions in red.}
\label{fig:heatmap_grid}

\end{figure*}

\Cref{tab:mAP} reports detection performance across different lesion sizes (as defined in \cite{mourao2025multi}). MIL-PF achieves competitive results for medium and large sizes, while mAP for small lesions remains low. However, qualitative results on randomly selected examples in \cref{fig:heatmap_grid} show that our models mostly struggle with reaching IoU@0.25 threshold, while consistently identifying the correct principal regions. We attribute this to the relatively large input tile sizes ($448 \times 448$ / $518 \times 518$), which, while not hindering leading classification performance, affect detection resolution. We consider this a possibility for future enhancement.

\begin{table}[!ht]
\centering
%\caption{Quantitative comparison of proposed MIL-PF \wrt SOTA models. On the more complete and challenging dataset EMBED, MIL-PF shows superior performance than all previous methods.}
\caption{Comparison of proposed MIL-PF with FPN\cite{mourao2025multi}. Detection performance is reported for all (mAP), small (mAP$_s$), medium (mAP$_m$) and large (mAP$_l$) lesions.}
{\resizebox{\columnwidth}{!}{
\begin{tabular}{lcccccccc}
\toprule
& \multicolumn{4}{c}{\textbf{Calcification}}  & \multicolumn{4}{c}{\textbf{Mass}} \\ 
\cmidrule(lr){2-5}\cmidrule(lr){6-9}
\textbf{Model/Paper} & \textbf{mAP} $\uparrow$ &  \textbf{mAP$_s$} $\uparrow$ & \textbf{mAP$_m$} $\uparrow$ & \textbf{mAP$_l$} $\uparrow$ &  \textbf{mAP} $\uparrow$ &  \textbf{mAP$_s$} $\uparrow$ & \textbf{mAP$_m$} $\uparrow$ & \textbf{mAP$_l$} $\uparrow$ \\ 
\midrule

FPN-AbMIL \cite{mourao2025multi}                        & 32.0 & 9.1 & 34.8 & 57.5 & \textbf{28.2} & \textbf{4.7} & \textbf{32.1} & 66.2\\
\rowcolor{gray!10}FPN-SetTrans \cite{mourao2025multi}   & 37.4 & \textbf{18.8} & 39.5 & 62.2 & 24.3 & 3.0 & 28.0 & \textbf{73.2}\\
\textbf{MIL-PF (Ours, DINOv2)}                          & 24.4 & 0.4 & 22.5 & \textbf{94.44} & 16.3 & 0.1 & 15.9 & 62.8 \\
\rowcolor{gray!10}\textbf{MIL-PF (Ours, MedSigLIP)}     & \textbf{38.4} & 1.2 &\textbf{ 52.5 }& 84.9 & 14.1 & 0.1 & 11.3 & 58.4 \\
\bottomrule
\end{tabular}
}}
\label{tab:mAP}
\end{table}

\subsection{Ablation Studies}
\label{sec:ablation}
To quantify the significance of each component of MIL-PF, we train differently configured MIL architectures. The results are shown in \cref{tab:two_tables}.

\begin{table}[t]
\centering
\caption{Encoder selection (left) and ablation on aggregation types (right) on EMBED.}

\begin{minipage}{0.49\linewidth}
\centering
\resizebox{\linewidth}{!}{
\begin{tabular}{ccccc}
\toprule

\textbf{Model} & \textbf{Resolution} & \textbf{AUC} $\uparrow$& \textbf{Spec@Sens=0.9} $\uparrow$\\ \midrule

% SIL$^{IL}$ \cite{pathak2025breast}  & $2944\times1920$ & 0.816 & 0.389 \\
MammoCLIP~\cite{ghosh2024mammo}  & $1520\times912$ & 0.870 & 0.558 \\
\rowcolor{gray!10}BiomedCLIP~\cite{zhang2023biomedclip}  & $224\times224$ & 0.872 & 0.606\\
MedSigLIP~\cite{sellergren2025medgemma}  & $448\times448$ & \textbf{0.897} & \textbf{0.691}\\
\rowcolor{gray!10}RADDINO~\cite{perez2024rad} & $518\times518$ & 0.854  & 0.499\\
DINOv2~\cite{oquab2023dinov2} ViT Giant  & $518\times518$ & \textbf{0.897} & \underline{0.655}\\
\rowcolor{gray!10}DINOv3~\cite{simeoni2025dinov3} ViT Huge+  & $512\times512$ & 0.831 & 0.497 \\
DINOv3~\cite{simeoni2025dinov3} ViT Huge+  & $1024\times1024$ & 0.834 & 0.476 \\
\bottomrule

\end{tabular}
}
\end{minipage}
\hfill
\begin{minipage}{0.49\linewidth}
\centering
\resizebox{\linewidth}{!}{
\begin{tabular}{llcccc} % Changed from lcccccc to llcccc (6 columns)
\toprule
% New header structure spans 2 columns for Aggregation
\multicolumn{2}{c}{\textbf{Aggregation}} & \multicolumn{2}{c}{\textbf{DINOv2 ViT Giant}} & \multicolumn{2}{c}{\textbf{MedSigLIP}} \\
\cmidrule(lr){1-2} \cmidrule(lr){3-4} \cmidrule(lr){5-6} % Adjusted cmidrules for new columns
% Specific column headers
\textbf{Global ($\mathcal{A}^G$)} & \textbf{Local ($\mathcal{A}^L$)} & \textbf{AUC} $\uparrow$ & \textbf{Spec@Sens=0.9} $\uparrow$ & \textbf{AUC} $\uparrow$ & \textbf{Spec@Sens=0.9} $\uparrow$ \\
\midrule
% Data rows with aggregation types split into two columns
mean (inf. only) & --- & 0.871 & 0.627 & 0.870 & 0.619 \\
\rowcolor{gray!10}mean & --- & 0.878 & 0.640 & 0.873 & 0.640 \\
max (inf. only) & --- & 0.865 & 0.601 & 0.873 & 0.621 \\
\rowcolor{gray!10}max & --- & 0.897 & 0.665 & 0.897 & 0.691 \\
attn & --- & 0.896 & 0.668 & 0.899 & 0.676 \\
\rowcolor{gray!10}--- & attn & 0.892 & 0.662 & 0.899 & 0.713\\
max & mean & 0.901 & 0.705 & 0.902 & 0.672 \\
\rowcolor{gray!10}max & max & 0.905 & 0.703 & \textbf{0.918} & 0.735 \\
max & attn & \underline{0.916} & \textbf{0.762} & 0.914 & \underline{0.746} \\
\bottomrule
\end{tabular}
}
\end{minipage}

\label{tab:two_tables}
\end{table}

The configurations containing $\mathcal{A}^G$ use global features, the ones containing $\mathcal{A}^T$ use tile features, and the ones containing both, use both, as in full MIL-PF.
The aggregations labeled with \textit{inf. only} are only performed at the inference level, while for the training each image inherits its bag's label. This simulates non-MIL approaches, but aggregation is still preserved at inference to preserve consistency.

We note that the difference between the full MIL-PF setup and na\"ive single instance learning (SIL) performed only on global views goes up to 5\% for AUC and 14\% for Spec@Sens=0.9, proving MIL-PF's inductive biases correct and its capacity capable of leveraging powerful backbones.

\section{Conclusion and Future Work}
\label{sec:conclusions}

We introduced MIL-PF, a lightweight design for weakly-labeled mammography data that operates on features precomputed by frozen foundation models. Our key finding is that a powerful general-purpose encoder, combined with a small specialized $\sim$40k parameter task-specific head, yields state-of-the-art results on a large-scale clinical dataset. Results challenge the prevailing assumption that end-to-end fine-tuning or complex custom architectures are prerequisites for this task. This opens multiple venues for future research, including the integration of more complex inductive biases, such as prior patient exams or bilateral asymmetry impact, and the application to other high-resolution, weakly-labeled domains.

\bibliographystyle{splncs04}
\bibliography{main}

\end{document}